# A Data-Center FPGA Acceleration Platform for Convolutional Neural Networks

Xiaoyu Yu, Yuwei Wang, Jie Miao, Heng Zhang, Yu Meng,
Bo Zhang, Biao Min, Dewei Chen, Jianlin Gao
*Tencent Shenzhen, China*
{kevinxiaoyu, derickwang, austingao}@tencent.com

Ephrem Wu,
*Xilinx, Inc., San Jose, CA 95124, USA*
ephrem.wu@xilinx.com

*Abstract*—Intensive computation is entering data centers with multiple workloads of deep learning. To balance the compute efficiency, performance, and total cost of ownership (TCO), the use of a field-programmable gate array (FPGA) with reconfigurable logic provides an acceptable acceleration capacity and is compatible with diverse computation-sensitive tasks in the cloud. In this paper, we develop an FPGA acceleration platform that leverages a unified framework architecture for general-purpose convolutional neural network (CNN) inference acceleration at a data center. To overcome the computation bound, 4,096 DSPs are assembled and shaped as supertile units (SUs) for different types of convolution, which provide up to 4.2 TOP/s 16-bit fixed-point performance at 500 MHz. The interleaved-task-dispatching method is proposed to map the computation across the SUs, and the memory bound is solved by a dispatching-assembling buffering model and broadcast caches. For various non-convolution operators, a filter processing unit is designed for general-purpose filter-like/pointwise operators. In the experiment, the performances of CNN models running on server-class CPUs, a GPU, and an FPGA are compared. The results show that our design achieves the best FPGA peak performance and a throughput at the same level as that of the state-of-the-art GPU in data centers, with more than 50 times lower latency.

## I. INTRODUCTION

Developing a hardware computing platform for convolutional neural network (CNN) based deep learning inference in a modern data center is a significant challenge [1–2]. The introduction of AlexNet has enabled the creation of very deep CNNs with hundreds of layers that improve the accuracy [3–5]. Subsequently, most research has focused on reducing the redundancy and improving the efficiency to lower the computational cost in applications, such as low-bit representation [6–8], parameter pruning and sparsification [9], Winograd/Fast Fourier Transform (FFT)-based optimization, and implementation of novel layer types such as squeeze and shuffle [10]. Hence, in terms of the total cost of ownership (TCO), a reconfigurable accelerator with homogeneity on the hardware level is desirable, particularly when models are still in fast evolution. The field-programmable gate array (FPGA) is an ideal choice for maintaining the same infrastructure and provides customized computing architectures for different solutions. However, three aspects of the FPGA architectures for online CNN inference still need to be discussed: achieving a higher throughput to lower the inference cost per image in CPU-FPGA-based servers, efficiently supporting diverse CNN workloads, which show quite different input image sizes, model topologies and basic operators, and exploring architectures that could be transplanted from one model to another from hours to minutes.

In this paper, an FPGA acceleration platform based on supertile methods is proposed for general-purpose CNNs in a data center. The principal contributions are as follows:

1. A unified and scalable framework is proposed for a CNN accelerator for applications in a data center, in which basic supertile units (SUs) are scaled up with an interleaved task dispatch to achieve maximum performance and efficiently support different types of convolution.

2. A dispatching-assembling buffering model with broadcast cache (BC) sets is designed for a multi-SU architecture and to scale up the reading and writing bandwidth.

3. Postprocessing architectures with logic sharing are proposed to support various operations and simplify the design. A two-dimensional filter processing unit (FPU) for a class of filter-like and pointwise operations is discussed to balance design complexity and performance.

The remainder of this paper is organized as follows: Section II discusses the related works. Section III provides the supertile-based design for the convolution computation. Section IV focuses on the design for non-convolution operators. Section V discusses the memory organization and implementation of the system. After a comparison with CPUs, a GPU and related works in Section VI, Section VII concludes the paper.

## II. RELATED WORKS

FPGAs have been adopted by most cloud service providers, such as Amazon, Microsoft, Tencent, Baidu, Alibaba, and Huawei, as a reconfigurable heterogeneous computing resource. NN-based inference solutions on an FPGA have also been discussed in [8, 11–16]. Project Catapult and Brainwave from Microsoft are the most widely deployed examples of FPGAs in data centers on both the infrastructure and application levels [14] for search ranking, network acceleration low-latency LSTM [11], and CNN processing [15, 16]. The solutions also provide interconnection across chips, cards, and servers and organize FPGAs at the data center scale into an acceleration pool. Baidu [12] developed a software-defined accelerator for matrix multiplication on an FPGA, and the active functions were reconfigurable depending upon different cases to fit more models.

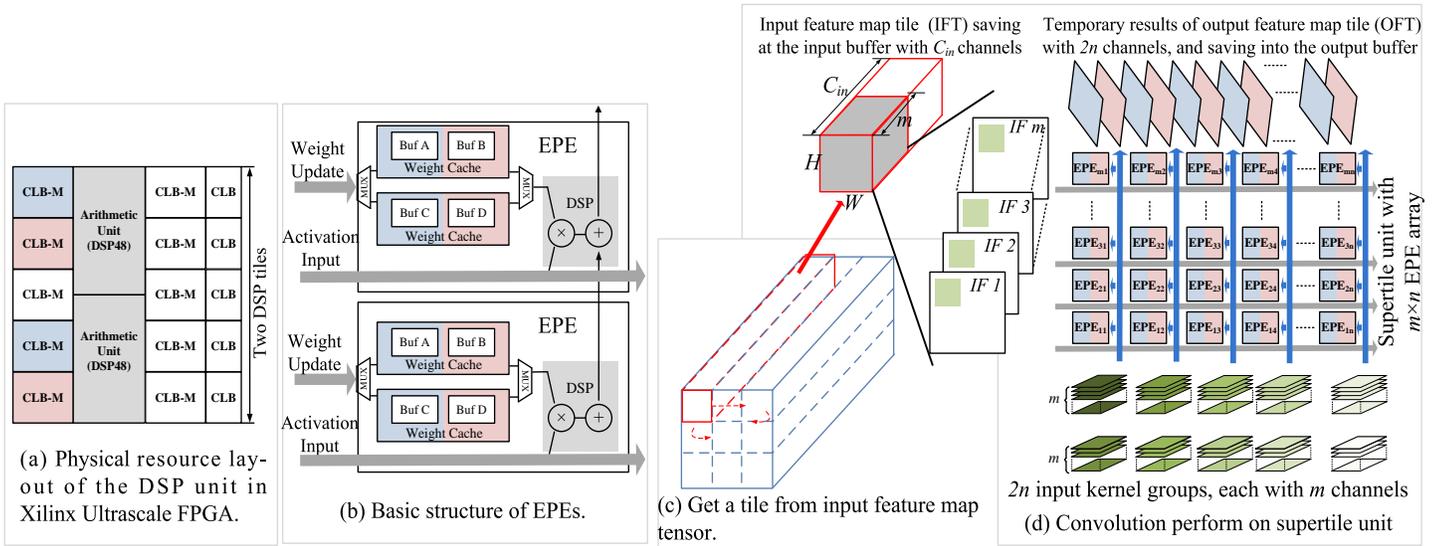

Fig. 1. Convolution with an input feature map tile (IFT) and 2n kernel groups on one SU.

In the recent literature, studies have proposed CNN accelerators on high-end FPGAs that enhance the processing abilities of inference for cloud services. [17] and [18] used an OpenGL-designed architecture to accelerate AlexNet and VGG on Arria 10. A reusable CNN engine with a unified framework and a scalable PE array was proposed in [19], which provided an end-to-end solution for deploying CNN models from Caffe onto an FPGA. The motivation matches well with the gap between deep learning researchers and hardware, but there is still space to improve the performance and resource utilization. [20] proposed a layer-based pipeline structure built by automated tools for both edge and cloud applications, which provided a deeply optimized architecture for different models.

Previous solutions on high-end FPGAs have the following drawbacks:

1) Limited architecture scalability: A number of studies have discussed solutions on middle/low-end FPGAs with dedicated off-chip memory. When on-chip resources, such as digital signal processors (DSPs), change from hundreds to thousands and the off-chip bandwidth remains, the scalability of the processing elements and memory system has not been sufficiently discussed.

2) Limited types of operators: Various kinds of convolution and non-convolution operators are emerging in CNN-model design, and only a few types of operators have been given significant attention in FPGA-based architecture design.

3) Higher deployment cost: Instead of a specific model, inference tasks involve many types of CNN networks. Deploying a new model onto an existing CNN architecture without general-purpose orientation design would cost extra development time. Although exploration methods with automatic tools may perform well for bottleneck analysis and resource scheduling, the time cost associated with re-synthesis in the deployment remains.

In this paper, we focus on the drawbacks above and discuss a scalable solution both for the computation and memory architecture on high-end FPGAs and reduce the deployment cost for different models with a general-purpose design.

### III. UNIFIED COMPUTING ENGINE FOR CONVOLUTION

Like multi-pumping used in [21], the supertile method proposed in [22] runs the DSP systolic array at twice the clock rate of the surrounding system logic. This approach has three benefits: 1) extensive use of built-in DSP cascades enables the systolic array to operate at maximum throughput while consuming little fabric resources; 2) the DSP is efficiently used as both a multiplier and an adder; and 3) the same input data is reused and multiplied by at least two different weights from the local weight buffer in each DSP supertile. However, [22] focused on the supertile model and computing behavior from the DSP to a 2D processing array. Cross-array processing, task mapping and scheduling, various types of convolution adaption, and on-chip memory design are still under discussion. In this section, we focus on the challenges of scaling up a supertile unit (SU) to multiple units and employing different types of convolution.

### A. Supertile Unit

Fig. 1 (b) shows the structure of the enhanced processing element (EPE), which is designed to fit the physical resource layout of the Xilinx Ultrascale FPGA in Fig. 1 (a) for better timing performance. An EPE running at the double clock rate would increase the bandwidth demand both for activations and weights. Therefore, in each EPE, small distributed RAMs cache weights in ping-pong mode for fast responses. When one of the weight caches supplies weights for computing, the other waits for weight update to overlap the weight transfer with the computation. The weight cache also contains ping-pong buffers running at twice the clock speed to provide weight data to the DSP. The activation input is not only shared in the local EPE but also shared with all EPEs within the same row. When the EPEs are spread in the form of an $m \times n$ array, this array becomes an SU as shown in Figs. 1(c) and (d).

When the data from a $k_x \times k_y$ sliding window of a feature map are streamed into a row of EPEs as a 1D vector, the corresponding weights are fetched from the weight caches for dot-product operations. The weight-activation products in the same position of the sliding window from different channels in the EPEs along the same column are summed and then produced at the top of each column. Because EPEs run at the double clock rate, two results from two kernel groups are stored into two output buffer blocks at every $k_x \times k_y$ cycles.

*B. Scaled-up SU*

We organize the DSPs into two levels, with the first level being the SU and the second level crossing the SUs. Different from [22], we set $n = 16$ and $m = 32$ for each SU containing 512 DSPs for two reasons: For a model-based consideration, the number of channels of most input and output feature maps is a multiple of 32 in [3-5, 25-26, 28], and for the hardware design, the input and output data paths prefer a matched data bit width in the loop. Two challenges exist when putting SUs into practice. First, directly deploying individual batch-based tasks onto SUs would introduce more resource cost both in terms of memory and bandwidth. A proper method for the task partition should be explored to fully use SUs simultaneously and efficiently. Second, the input and output data bandwidth would be multiplied when multiple SUs are applied. Dedicated design for data buffering should be discussed. We propose the interleaved task dispatching method and a dispatching-assembling buffer model to solve these problems, as shown in Fig. 2.

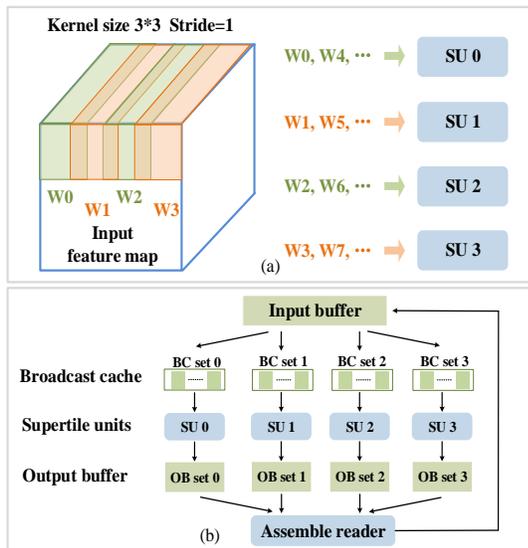

Fig. 2. (a) Interleaved task dispatch based on sliding windows to the SUs and (b) dispatching-assembling buffering model with the BC.

Interleaved task dispatching explores column-based parallelism for fine-grained task partitioning, as shown in Fig. 2(a). The same *2n* kernel groups are loaded into four SUs for weight sharing before the convolution. Initially, the vectors from successive sliding windows are sent to four SUs. The activation from the same IFT feeds into the SUs, but the window locations are different. The convolutions of the four sliding windows are processed simultaneously, which provides parallelism in one row's processing.

Each SU calls for its exclusive reading and writing memory bandwidth, and a dispatching-assembling buffer model is proposed, as shown in Fig. 2(b). With the case of four processing paths, the data buffered in the input buffer (IB) are shared among the paths. Each path contains a broadcast cache (BC) set as the local data cache, an SU as the processing unit, and an output buffer (OB) set providing writing bandwidth and buffering temporary convolution results. Under the control of interleaved task dispatching, the four paths work synchronously, focusing on the same convolutional task. When the output tensor is ready after convolution, an assemble reader reads the data distributing across the OB sets, reorders the data, and writes back to the IB for the next convolution. Note that the IB provides only one-fourth of the total input bandwidth for the SUs. To meet this discrepancy, multi-BC sets are used to buffer the same temporary tensor dispatched from the IB and output the data from different sliding windows for the SUs, which are further discussed in Section V.

In this way, sliding-window-based tasks are dispatched onto multiple SUs, and the memory bottleneck in data reading and writing is avoided. Finally, up to 4096 DSPs are organized as 8 SUs and are shaped as 3D tensor processing units of 2 CNN engines on 2 die of KU115. The four SUs in each engine share the same IB, kernel load controller, and kernel data, thus allowing the deployment of more DSPs on one convolution layer and reducing the time cost of processing by a factor of four.

*C. Tile-Based Slice-Loop-Hiding Cross Input Feature Map*

The definition of the tile is a sub tensor, with a smaller size in *H* and *W* but the same size in $C_{in}$ compared with the original tensor (Fig. 1 [C]). The slice is a sub tensor with the same *H* and *W* but that contains only a few of channels on $C_{in}$, like the IFT with *m* channels in Fig. 1 (d). Instead of separated slice processing with respective commands, a slice-loop controller is placed between the command decoder and SUs' controller, handshaking with them, controlling the slice loop and updating a few address parameters automatically with the following benefits: 1) The on-chip memory is used efficiently, and no extra storage capacity would be introduced to buffer the results between slices. 2) The command efficiency is improved and the total command length, command loading and decoding times are reduced. 3) The idle time of the SUs are minimized, and the next-slice processing is triggered immediately in the local controller instead of via talking and handshaking with the global controller.

*D. Efficient Processing for Different Types of Convolutions on SUs*

A standard convolution with different parameters, together with various types of convolutions, appears in CNN-model evaluation, wherein the utilization and real performance vary significantly [20, 34]. We will discuss methods to map the convolutions onto this unified SU-based architecture efficiently, which include the following special cases.

**The first layer of a standard convolution** has only three channels, and only $3/m$ DSP-rows of each SU can be used without optimization. We partition the data within a sliding window of a channel into pieces and send them into different

EPE rows with the help of the parameters *window_pos_begin* and *window_pos_end*. Thereafter, the partition results of a sliding window are added together with cascade adders between the EPEs in a column. The number of pieces (*NP*) is defined as $NP = \max\{ks, ks \times 3, ks \times ks\}$, where $NP \leq m$. For example, when $m = 32$ and the kernel size (*ks*) of the first layer is $7 \times 7$ with 3 channels, the computation proceeds as $ks = 1 \times 7$ with 21 channels to improve the utilization of the SUs from 3/32 into 21/32.

**Kernel fusion for the non–first layer of a standard convolution with $ks = 1$**: A convolution with $1 \times 1$ kernels usually accompanies with lower reusability of activation and changing bound from computation to memory. In contrast to the kernel partition, kernel fusion is applied. When the kernel group number is $C_{out}$ and when $Fu = \lceil C_{out}/2n \rceil$, where $Fu \in [1,16]$, each weight buffer with a depth of 16 can buffer the weight with the number of *Fu*. For example, when $n = 16$ and when the kernel tensor with the size $1 \times 1 \times C_{in} \times 384$ is convolved with an input tile tensor, e.g., $Fu = \lceil \frac{384}{32} \rceil = 12$, weight buffer A at EPE$_{ij}$ in Fig. 1 (d) caches 12 kernels with the indices $\{C_{in} = i, C_{out} = j + Sn \times 32\}$, where *Sn* is the slice index with a value of 0, 1,..., 11. When processing is enabled, the activation sent into the SUs is updated every 12 clock cycles and shared with 12 weights in each weight buffer at each EPE. After accumulation with the cascaded adder, the data at the end of each EPE column are updated every clock cycle and written into the OB with the corresponding OFT address.

**Nonstandard convolution types** comprise the transposed convolution [35], dilated convolution [36], and depthwise convolution [27]. Both transposed and dilated convolutions can be computed as a standard convolution layer, except up-sampling should be performed before the transposed convolution. Depthwise convolution is performed in an FPU, which will be discussed in Section IV *A*.

## IV. POSTPROCESSING DESIGN

The convolution consumes most of the computation in the CNN model with only a few types, whereas non-convolution operations have low computational costs and comprise most types of operators in the opposite manner in different models (e.g., pool, normalization, activation function, and element-wise operations between branches [23-26].). We divide these operations into three classes: 2D filter-like operations with the sliding window traverse height (H) and width (W) of a channel of the feature map, operations across channels (C) and operations that can fuse with convolution to reduce the memory access. We design a general-purpose processing unit for the first category and a custom module for operators (such as LRN) in the second category because they do not frequently appear in current inference tasks. The operations in the third category are processed with operator fusion.

### A. 2D Processing Unit for Filter-Like Operations

When designing the circuit for 2D filter-like operations, we consider the following aspects: The first is compatibility and configurability to support the current and potential operators. Simplifying the complexity of the hardware is the second consideration to avoid multi-module scheduling, maintain a large parameter field, and provide a dedicated data load and storage path with a complex multiplexor for each module. The third aspect is the resource limitations. SUs consume most of the resources of a specific physical region after mapping, as shown in Figs. 1 (a) and (b), and only a few resources outside the region are available. The resource limitation calls for more functional logic sharing. In this case, an FPU is proposed.

We observe that there are common computing and data access styles in filter-like non-convolution operators that traverse the feature map within a sliding window (kernel) on each channel, and no operations exist across channels, such as max/average pool. Most of the parameters also have similarities, such as the parameters of the source address reading like *StartAddr/Stride/Pad/WindowSize/FeatureMapSize/ChannelNum* and the parameters of destination address writing like *StartAddr* and *FeatureMapSize*. Pointwise operations, such as relu/relu6/linear-transformations, can be considered special cases when the kernel size is equal to 1. A depthwise convolution [27] could also be performed on the proposed FPU because no cross-channel addition is needed. Better performance can be achieved when standard and depthwise convolutions are interweaved [27–28], which can be performed with the SUs and FPU, respectively, in parallel.

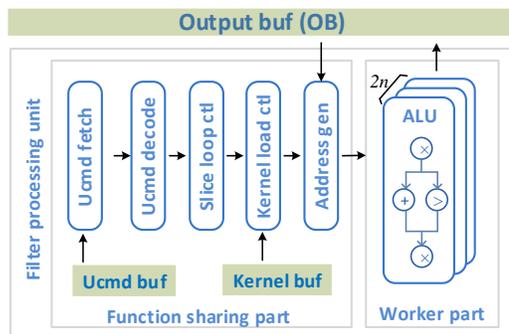

Fig. 3. Architecture of the FPU.

Fig. 3 shows the micro-architecture of the FPU. It is mainly divided into the function sharing part (FSP), which contains the common functional modules in the processing for different operators, and the worker part (WP), which can be changed or integrated into more processing modules depending on the needs. The FSP fetches the micro-commands from the ucmd buffer [37], decodes them into parameters including the parameter categories of DataLoad/DataStore/FuncSet/Scalar-Value. Then, the tensor are read by the address generation module from OB continuously, streamed into WP for processing, and finally written back into the OB. When the channel number is more than *2n*, the slice-loop-control module is enabled for better command efficiency as slice-loop-hiding for convolution. In the WP, 2n channels of ALUs process the data stream in SIMD mode. The reconfigurable ALU is designed with a cascaded pre-multiplier, mid-adder/comparator, and final multiplier, which can be enabled or bypassed depending on the function setting by the micro-command. To improve the efficiency and save memory bandwidth, they work in pipelines and provide a maximum of three operations per clock. Note that the module for the kernel load control is located next to slice-loop controller, providing the weights when depthwise convolution is enabled. For max-pool, the mid-comparator is enabled, and the others are bypassed. In mid-comparator, the maximum value is saved into the register after each comparison, and the *window-end* signal triggers the output and resets the register. The final multiplier is used for the division of avg-pool, and the adder is also enabled in this case.

*B. Operator Fusion*

Each operation across the tensor would introduce extra memory access, which would suspend other memory-access-dependent operations. This is the reason we fuse some pointwise operators with the convolution even though they have already been supported by the FPU. Fig. 4 shows a cascade of four kinds of operations at the end of the column of each SU. The first adder adds the output of the SU with the temporary results of the OB from the previous slice. The second adder is for element-wise addition across the branches, which is widely used as a residual block [23]. Relu is then performed, and dynamic precision data quantization [7] follows. Finally, the results are written into the OB. These operations can be individually enabled or bypassed with control instructions.

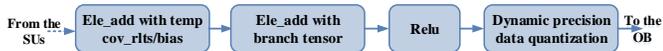

Fig. 4. Operator fusion in postprocessing.

## V. IMPLEMENTATION OF THE ACCELERATOR

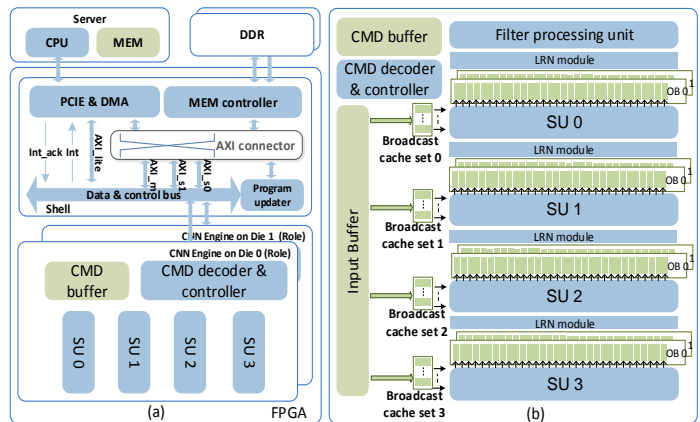

Fig. 5 System implementation: (a) the overall system and (b) CNN engine in one die.

*A. System Overview*

Fig. 5 shows a heterogeneous server architecture with the CPU + FPGA, including a server system with the CPU and memory, two channels of DDR4 on an FPGA-PCIE card, and two CNN engines in the FPGA. The memory/buffer is shown in green, and the control/processing logic is shown in blue. Here, 2048 DSPs shaped as four SUs running at the double clock speed provide enough computing capacity in each engine, while FPU and operator fusion at the output of each SU integrate most of the non-convolution operators and simplify the design. Only the modules customized for LRN remain.

*B. Memory Organization*

The on-chip memory is mainly divided into IB and OB sets as shown in Fig. 5(b). The IB is shared globally among multiple SUs. The OB sets are placed at the output of each SU and each set consists of 2n components. Each component provides an exclusive read and write port for one SU's column. The OB sets are designed in the form of a ping-pong structure so that when n = 16, a 64 GB/s on-chip reading and writing bandwidth can be provided for both SUs and the FPU, allowing them to run in parallel to overlap the convolution and filter-like operations.

*C. Broadcast Cache*

When four SUs operate together, each consumes the input bandwidth of $16 \times 32 \times f_{logic}$ bit/s in Fig. 2(b), where $f_{logic}$ is the frequency of non-EPE logic, and up to $4 \times 512 \times f_{logic}$ bit/s of bandwidth is needed. However, the output bandwidth of the IB is only $512 \times f_{logic}$ bit/s. The BC is designed to meet this discrepancy and buffer the same input feature maps but output the data from different sliding windows for column-based parallelism. The BC at the input port of each row of the SU is a circular buffer that updates the data continuously. The data used will be overwritten with the data from the next row. The window in each BC continues to slide with a step size of 4 × *Convolution-Sride* along the row but starts at different positions for different SUs. Fig. 6 shows the behavior of one BC with a $3 \times 3$ sliding window inside and with *Convolution-Stride*=1.

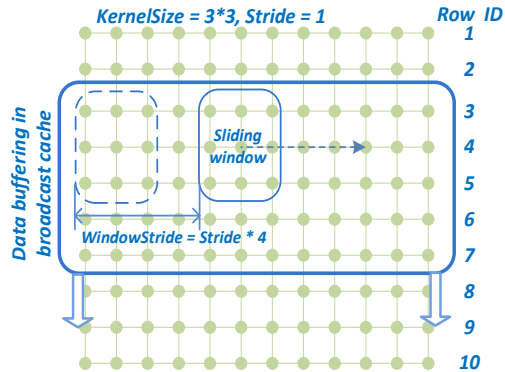

Fig. 6. The behavior of the BC for sliding window sequence sampling.

In this case, the data within *rows 3* to *7* are buffered in the cache, and the window moves with the central position on *row 4* and *Window-Stride = 4*. After the computation for the last sliding window position of *row 4*, the window center moves to the first position of *row 5*, and the BC starts to load the data of *row 8* to overwrite the data of *row 3*. To accommodate enough rows to fit the sliding window and leave an extra row as the margin, the cache should buffer at least *Kernel-Size + 1* rows of data. On the basis of the tile partition method in Section III *C*, the width of the tile of the input feature map can be flexibly narrowed to buffer more rows when a larger kernel size is used.

## VI. EXPERIMENT AND PERFORMANCE

The performance of the implemented platform is evaluated in this section. We setup the system on the server and perform three CNN models on it. After that, the performance of the proposed system is compared with those of the other CNN acceleration solutions in the data center, including high-end FPGAs, CPUs, and a GPU.

*A. Experimental Setup*

The proposed CNN engines are implemented on KU115 with Vivado 2016.4. Table I lists the resource utilization for AlexNet/GoogLeNet. Each type of resource exceeds 70% of the total, thus making it difficult to reach the maximum frequency of 661 MHz in [22]. Finally, at the peak performance of 4.2 TOP/s with 16-bit quantization, 500 MHz is used for the EPEs, and 250 MHz is used for the others. The system is built on Semptian's FPGA card with a PCIe interface, and the size of the card is half height and half length (Fig. 7, left side). SUPERMICRO 6028UX-TR4 (Fig. 7, right side) is used as a server with two Intel Xeon E5-2680V4 CPUs and 16 GB ×16 of DDR3 SDRAM.

TABLE I. FPGA RESOURCE UTILIZATION

|  | LUT | FF | BRAM Blocks | DSP |
|---|---|---|---|---|
| **Used** | 469091 | 967577 | 1540 | 4214 |
| **Available** | 663360 | 1326720 | 2160 | 5520 |
| **Utilization** | 70.7% | 72.9% | 71.3% | 76.3% |

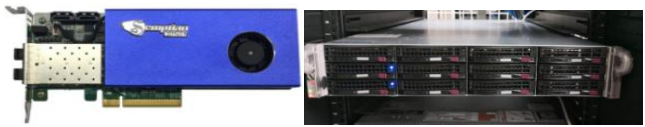

Fig. 7. Hardware platform of the experiment.

*B. Acceleration for Different CNN Models*

The experiments are performed with three models (Table II). AlexNet is well discussed in most of the literature on accelerator design, and 92% and 8% of the computations are performed on the FPGA and CPU, respectively. The performance reaches 2.3 TOP/s with a latency of 2.3 ms. The second model is GoogLeNet. We adjust the batch size to 2 and achieve approximately 1.6 TOP/s with a 3.8 ms latency. We select the high-concurrency network (HCNet) as the third model, which is a customized compact CNN model to lower the classification cost per image at Tencent. This model achieves almost the same accuracy as GoogLeNet but three times the throughput upon testing on an Intel Xeon E5-2620v3. Inspired by ResNet [5] and ShuffleNet [10], the HCNet begins with the convolution and pooling layers changing the input image from 224×224 into 56×56 with 32 channels. Three stages follow until the end of the model with avg-pooling and FC layers. There are four, eight, and four basic residual blocks (see Fig. 8) in three stages, respectively.

The 1 × 1 convolution and fewer channel convolutions are largely used in the HCNet. Although kernel fusion for 1×1 is performed, fewer output channels decrease the reusability of the activation, and frequent data transfer lowers the SU utilization, which increases the power cost. Furthermore, when one-tenth of the layers are those with 16 channels, the SUs are not fully used. Although a performance of 3.44× is achieved compared with P4 with a 7 ms time constraint [2], its throughput is limited to 650.5 GOP/s at 225/450 MHz. A higher frequency and platform with the same design as AlexNet/GoogLeNet could be used if the limitation on the PCIE power supply could be ignored. Owing to the efficient model structure, 2.8 times the throughput of GoogLeNet is achieved, and the TCO is significantly reduced. By comparing the performance with Nvidia TESLA P4, which is the state-of-the-art GPU for deep learning inference in data centers, the speedup ratios of the FPGA in these three tests are 1.35, 3.91 and 3.44, respectively, with a 7 ms response time limitation.

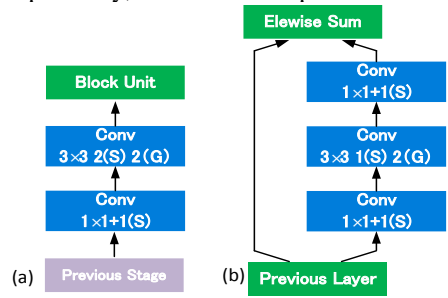

Fig. 8. The residual blocks shown in (b) are used as the basic building block for the HCNet. Each block consists of three convolutional layers, where two 1 × 1 convolutional layers are used for feature squeezing and unsqueezing, and a 3 ×3 convolutional layer with group 2 is used for spatial convolution. Down-sampling is performed at the beginning of each stage, where the 3 × 3 convolution with stride 2 is used, as in (a). The kernel group numbers of these layers in three stages are [32, 32, 128], [64, 64, 256], [128, 128, 512].

TABLE II ACCELERATION FOR THREE CNN MODELS.

|  | **AlexNet** | **GoogLeNet** | **HCNet** |
|---|---|---|---|
| **Data precision** | 16-bit | 16-bit | 16-bit |
| **Clock (MHz)** | 250/500 | 250/500 | 225/450 |
| **Batch size** | 4 | 2 | 4 |
| **CNN size (MOPs)** | 1331.6/1448.8 | 3081.0/3083.1 | 444 |
| **Throughput (FPS)** | 1753.8 | 527.7 | 1465.1 |
| **Performance (GOP/s)** | 2335.4 | 1625.9 | 650.5 |
| **Latency (ms)** | 2.3 | 3.8 | 2.7 |
| **Power (watts)** | 62.6 | 56.6 | 57.6 |
| **Speedup VS P4 (7 ms)** | 1.35 | 3.91 | 3.44 |
| **Energy efficiency (GOP/s/W)** | 37.3 | 28.7 | 11.3 |

TABLE III. COMPARISON WITH FPGA-BASED CNN ACCELERATORS.

|  | **[18]** | **[19]** | **[20]** | **Ours** | |
|---|---|---|---|---|---|
| **FPGA chip** | Arria10-1150 | Virtex7-690t | KU115 | KU115 | KU115 |
| **Network** | VGG | AlexNet | VGG | GoogLeNet | AlexNet |
| **CNN size (GOPs)** | 30.8 | 1.4 | 30.8 | 3.1 | 1.3 |
| **Freq (MHz)** | 385 | 150 | 235 | 250/500 | 250/500 |
| **Precision** | Fix16 | Fix16 | Fix16 | Fix16 | Fix16 |
| **DSPs (used/total)** | 2756/3036 | 2833/3600 | 4318/5520 | 4214/5520 | 4214/5520 |
| **Peak performance (TOP/s)** | 2.1 | 0.8 | 2.1 | **4.2** | **4.2** |
| **Real performance (TOP/s)** | 1.79 | 0.6 | 2 | 1.63 | **2.3** |

## C. Comparison with FPGA-based Accelerators

Table III shows the comparison among different high-end FPGA-based CNN accelerators for potential cloud computing applications with the precision of fix16. We deliver the best peak performance of 4.2TOP/s, which is more than twice that of the others. In addition, real performance reaches 2.3 TOP/s in the AlexNet test benefiting from efficient task dispatching across multiple SUs, overlapping between computation and data-move, and minimizing the time cost in scheduling by slice-loop-hiding.

## D. Comparison with CPU and GPU

Table IV lists the available processing solutions that can be deployed in the data center for CNN inference. The Intel Xeon E5-2680V4 is a high-performance CPU and is also used as a host in the server for the FPGA/GPU. MKL 2018.0.0 is used for the optimization. The Nvidia TESLA P4 is a 16 nm GPU with 2560 CUDA cores and a 1 GHz clock speed and reaches 5.5 TeraFlops with the boosting of 192 GB/s memory bandwidth. CUDA 8.0.44 and Cudnn 6.0.21 are used to improve the performance. In the CPU test, two CPUs have 28 cores represented as 56 threads. Each thread binds with one of the batches with the corresponding batch size to keep all the cores busy. In the GPU test, a single process is used as a scheduler that sends the tasks to the GPU with different batch sizes. In the FPGA test, eight threads are used, and each binds with a physical core. Fig. 9 shows the results.

TABLE IV. BENCHMARKED SERVERS USING THE CPU, GPU, AND FPGA.

| Processor | Processor per server | TOP/s | | nm | MHz | On-chip memory (MB) | Off-chip memory BW (GB/s) | Power (Watts) | Release |
|---|---|---|---|---|---|---|---|---|---|
| | | *16bit* | *FP32* | | | | | | |
| Intel E5-2680V4 | 2 | - | - | 14 | 2400 | 35 ×2 | 76.8 ×2 | 120 | 2016 Q1 |
| NVIDIA P4 | 1 | - | 5.5 | 16 | 1000 | 10 [38] | 192 | 50-75 | 2016 Q3 |
| Xilinx KU115 | 1 | 4.2 | - | 20 | 250/500 | 11.8 | 38.4 | 50-66 | 2014 Q4 |

Running with a fixed batch size, the FPGA presents a steady performance, providing the lowest latency of 3.8 ms for GoogLeNet. The FPGA also shows the highest throughput until the batch size exceeds 32 for the GPU. Finally, the GPU achieves the highest throughput of 684 FPS at a batch size of 128 (it is out of memory at a batch size of 256). The situation is slightly different when the task moves to the HCNet. When the batch size ⩾ 4, the FPGA reaches its highest performance and maintains a constant batch size of 4. A comparison of the FPGA and GPU shows that the FPGA runs at a higher frame rate before the batch size of the GPU reaches 64. The gap does not change significantly until a batch size of 256 is reached. When P4 achieves its peak performance, the FPGA provides an 89% throughput with 1/57 latency compared with the GPU.

The performance of the FPGA is remarkable even when limitations exist. An FPGA with a simpler fabrication process, approximately 20% of the off-chip memory bandwidth, and one-fourth of the frequency of P4 can achieve superior performance in low-latency circumstances. For a larger batch size test, which higher data reuse is performed in the GPU, the performance improvement of the GPU over the FPGA is no more than 20%. The performance can be further improved when the proposed architecture is implemented with the next generation of FPGAs, e.g., UltraScale+ VU9P (16 nm), by using the same fabrication process of P4.

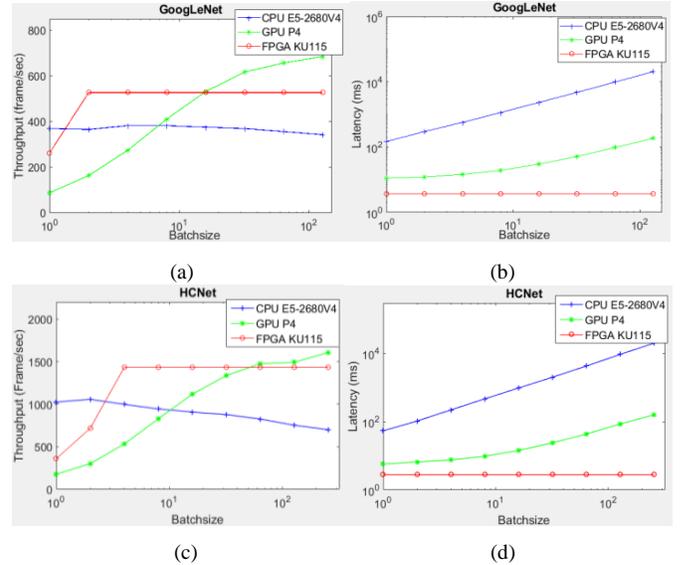

Fig. 9. Performance comparison with the CPU, GPU, and FPGA. (a) and (b) are the comparisons of the throughput and latency for GoogLeNet; (c) and (d) are the comparisons of the throughput and latency for HCNet. Note that the batch size in the FPGA is constant after the values of 2 and 4 for GoogLeNet and HCNet, respectively, because of the on-chip memory limitation.

## VII. CONCLUSION

In this paper, an FPGA acceleration platform with a supertile-based design is introduced for general-purpose CNNs and for performing various image/video inference tasks in data centers. Basic supertile EPEs are scaled up and shaped as multiple SUs to maximize the performance with the interleaved-task-dispatching method for the processing of types of convolution, and the increased bandwidth is provided by a dispatching-assembling buffering model. A configurable FPU is proposed due to the resource limitations to simplify the design and support different types of non-convolution operators, which makes it possible to run different CNN models on the same platform and reduce the deployment cost. We implement the design and make comparisons with high-end FPGAs and data-center-scale CPU/GPU. The experiment shows that the proposed architecture on KU115 achieves the best peak performance and throughput on FPGAs, and it performs at the same level as state-of-the-art GPU with more than 50 times lower latency. Compared with TCO, the FPGA enhances the throughput of the server by 149.2% with a 31.5% cost increase. The system is now deployed in a data center to serve over one billion people every day.

## ACKNOWLEDGEMENT

The authors acknowledge the people who contributed to our project and paper: Zhenyu Guo, Jianping Zhu, Qi Ju, Guanghui Wang, Jiaxi Li, Zhiqiang Cao, Zhuo Li, George Wang, Hu Zhao, Zexiong Ye and Jun Zhang.

# REFERENCES

[1] L. Ceze, M. D. Hill, and T. F. Wenisch. "Arch2030: A vision of computer architecture research over the next 15 years," arXiv preprint arXiv:1612.03182, 2016.

[2] N. P. Jouppi, et al. "In-data center performance analysis of a tensor processing unit," Computer Architecture (ISCA), 2017 ACM/IEEE 44th Annual International Symposium on. IEEE, pp.1–12, 2017.

[3] K. Simonyan and A. Zisserman. "Very deep convolutional networks for large-scale image recognition," arXiv preprint arXiv:1409.1556, 2014.

[4] C. Szegedy, et al. "Going deeper with convolutions," Proceedings of the IEEE Conference on Computer Vision and Pattern Recognition, 2015.

[5] K. He, X. Zhang, S. Ren, and J. Sun. "Deep residual learning for image recognition," In Proceedings of the IEEE Conference on Computer Vision and Pattern Recognition, 2016, pp.770–778.

[6] M. Rastegari, V. Ordonez, J. Redmon, and A. Farhadi. "Xnor-net: Imagenet classification using binary convolutional neural networks," European Conference on Computer Vision. Springer, Cham, 2016, pp. 525–542.

[7] J. Qiu, et al. "Going deeper with embedded FPGA platform for convolutional neural network," Proceedings of the 2016 ACM/SIGDA International Symposium on Field-Programmable Gate Arrays. ACM, 2016.

[8] Chung, E., Fowers, J., Ovtcharov, K., Papamichael, M., Caulfield, A., Massengil, T., ... & Boehn, C. (2017, August). Accelerating persistent neural networks at datacenter scale. Hot Chips (Vol. 27).

[9] S. Han, et al. "Ese: Efficient speech recognition engine with sparse lstm on Fpga," Proceedings of the 2017 ACM/SIGDA International Symposium on Field-Programmable Gate Arrays. ACM, 2017.

[10] S. Vivienne, Y.H. Chen, T.J. Yang, and J.S. Emer. "Efficient processing of deep neural networks: A tutorial and survey," Proceedings of the IEEE105.12 (2017), pp. 2295–2329.

[11] J. Fowers, et al. "A configurable cloud-scale DNN processor for real-time AI," 2018 ACM/IEEE 45th Annual International Symposium on Computer Architecture (ISCA). IEEE, 2018.

[12] J. Ouyang. "SDA: software-defined accelerator for large-scale deep learning system," VLSI Design, Automation and Test (VLSI-DAT), 2016 International Symposium on. IEEE, 2016, pp. 1–1.

[13] A. Putnam, et al. "A reconfigurable fabric for accelerating large-scale data center services," ACM SIGARCH Computer Architecture News, vol. 42, pp.13–24, 2014.

[14] A. M. Caulfield, et al. "A cloud-scale acceleration architecture," The 49th Annual IEEE/ACM International Symposium on Microarchitecture. IEEE Press, 2016, p. 7.

[15] K. Ovtcharov, O. Ruwase, J. Y. Kim, J. Fowers, K. Strauss, and E. S. Chung. "Accelerating deep convolutional neural networks using specialized hardware," Microsoft Research Whitepaper, vol. 2, 2015.

[16] K. Ovtcharov, O. Ruwase, J. Y. Kim, J. Fowers, K. Strauss, and E. S. Chung. "Toward accelerating deep learning at scale using specialized hardware in the data center," Hot Chips 27 Symposium (HCS), 2015 IEEE. IEEE, 2015, pp. 1–38.

[17] U. Aydonat, S. O'Connell, D. Capalija, A. C. Ling, and G. R. Chiu. "An OpenCL™ deep learning accelerator on arria 10," Proceedings of the 2017 ACM/SIGDA International Symposium on Field-Programmable Gate Arrays. ACM, 2017.

[18] J. Zhang, and J. Li. "Improving the performance of opencl-based fpga accelerator for convolutional neural network," Proceedings of the 2017 ACM/SIGDA International Symposium on Field-Programmable Gate Arrays. ACM, 2017.

[19] C. Zhang C, et al. "Caffeine: towards uniformed representation and acceleration for deep convolutional neural networks," Proceedings of the 35th International Conference on Computer-Aided Design. ACM, 2016, p. 12.

[20] X. Zhang, et al. "DNNBuilder: An automated tool for building high-performance DNN hardware accelerators for FPGAs," Proceedings of the International Conference on Computer-Aided Design. ACM, 2018.

[21] A. Canis, J. H. Anderson, and S. D. Brown. "Multi-pumping for resource reduction in FPGA high-level synthesis," Proceedings of the Conference on Design, Automation and Test in Europe. EDA Consortium, 2013.

[22] E. Wu, X. Zhang, D. Berman, and I. Cho. "A high-throughput reconfigurable processing array for neural networks," In Field Programmable Logic and Applications (FPL), 2017 27th International Conference on (pp. 1–4). IEEE.

[23] K. He, X. Zhang, S. Ren, and J. Sun. "Deep residual learning for image recognition," Proceedings of the IEEE Conference on Computer Vision and Pattern Recognition. 2016.

[24] C. Szegedy, et al. "Going deeper with convolutions," Proceedings of the IEEE conference on computer vision and pattern recognition. 2015.

[25] C. Szegedy, V. Vanhoucke, S. Ioffe, J. Shlens, and Z. Wojna "Rethinking the inception architecture for computer vision," Proceedings of the IEEE conference on computer vision and pattern recognition. 2016.

[26] C. Szegedy, S. Ioffe, V. Vanhoucke, and A. A. Alemi. "Inception-v4, inception-resnet and the impact of residual connections on learning," AAAI. vol. 4. 2017.

[27] F. Chollet. "Xception: Deep learning with depthwise separable convolutions," arXiv preprint (2017), pp. 1610–02357.

[28] A. G. Howard, et al. "Mobilenets: Efficient convolutional neural networks for mobile vision applications," arXiv preprint arXiv:1704.04861 (2017).

[29] M. Sandler, A. Howard, M. Zhu, A. Zhmoginov, and L. C. Chen. "Mobilenetv2: Inverted residuals and linear bottlenecks," 2018 IEEE/CVF Conference on Computer Vision and Pattern Recognition. IEEE, 2018.

[30] H. Li, X. Fan, L. Jiao, W. Cao, X. Zhou, and L. Wang. "A high performance FPGA-based accelerator for large-scale convolutional neural networks," Field Programmable Logic and Applications (FPL), 2016 26th International Conference on. IEEE, 2016.

[31] C. Zhang, P. Li, G. Sun, Y. Guan, B. Xiao, and J. Cong. "Optimizing FPGA-based accelerator design for deep convolutional neural networks," Proceedings of the 2015 ACM/SIGDA International Symposium on Field-Programmable Gate Arrays. ACM, 2015.

[32] R. Zhao, H.C. Ng, W. Luk, and X. Niu. "Towards Efficient Convolutional Neural Network for Domain-Specific Applications on FPGA," 2018 28th International Conference on Field Programmable Logic and Applications (FPL). IEEE, 2018.

[33] X. Zhang, X. Zhou, M. Lin, and J. Sun. (2017). "Shufflenet: An extremely efficient convolutional neural network for mobile devices," arXiv preprint arXiv:1707.01083.

[34] Y. Song, S. Liang, J. Wang, et al. The evolution of accelerators upon deep learning algorithms, Hot Chips 2018.

[35] A. Dosovitskiy, J. Tobias Springenberg, and T. Brox. "Learning to generate chairs with convolutional neural networks," Proceedings of the IEEE Conference on Computer Vision and Pattern Recognition. 2015.

[36] F. Yu, and V. Koltun. "Multi-scale context aggregation by dilated convolutions," arXiv:1511.07122, 2015.

[37] Moreau, Thierry, et al. "VTA: An Open Hardware-Software Stack for Deep Learning." arXiv:1807.04188, 2018.

[38] Jia, Zhe, et al. "Dissecting the NVIDIA Volta GPU Architecture via Microbenchmarking." arXiv:1804.06826, 2018.

[39] Chen, Yu-Hsin, Joel Emer, and Vivienne Sze. "Eyeriss: A spatial architecture for energy-efficient dataflow for convolutional neural networks." ACM SIGARCH Computer Architecture News. Vol. 44. No. 3. IEEE Press, 2016.